# A New CGAN Technique for Constrained Topology Design Optimization

M.-H. Herman Shen [1] and Liang Chen

Department of Mechanical and Aerospace Engineering

The Ohio State University

## Abstract

This paper presents a new conditional GAN (named convex relaxing CGAN or crCGAN) to replicate the conventional constrained topology optimization algorithms in an extremely effective and efficient process. The proposed crCGAN consists of a generator and a discriminator, both of which are deep convolutional neural networks (CNN) and the topology design constraint can be conditionally set to both the generator and discriminator. In order to improve the training efficiency and accuracy due to the dependency between the training images and the condition, a variety of crCGAN formulation are introduced to relax the non-convex design space. These new formulations were evaluated and validated via a series of comprehensive experiments. Moreover, a minibatch discrimination technique was introduced in the crCGAN training process to stabilize the convergence and avoid the mode collapse problems. Additional verifications were conducted using the state-of-the-art MNIST digits and CIFAR-10 images conditioned by class labels. The experimental evaluations clearly reveal that the new objective formulation with the minibatch discrimination training provides not only the accuracy but also the consistency of the designs.

## 1   Introduction

Topology optimization [1-4], a branch of design optimization, is a mathematical method to solve a material layout problem constrained for a given design domain, loading, and boundary conditions. This method determines the optimal distribution of material such that the structure has desired properties (e.g. minimizing the compliance of structure) while satisfying the design constraints. Indeed, topology design optimization offers a tremendous opportunity in design and manufacturing freedoms by designing and producing a part from the ground-up without a meaningful initial design as required by conventional shape design optimization approaches. Ideally, with adequate problem statements, to formulate and solve the topology design problem using a standard topology optimization process, such as Simplified Isotropic Material with Penalization (SIMP) is possible. However, the conventional optimization approach is in general impractical or computationally unachievable for real world applications due to over thousands of design iterations are often required for just a few design variables. There is, therefore, a need for a different approach that will be able to optimize the initial design topology effectively and rapidly.

---

[1] Corresponding Author, email: shen.1@osu.edu

There have been many studies to improve the topology optimization algorithms and reduce the computational costs using convolutional neural networks (CNN) [5-7]. The training process of the CNNs, mathematically, is an inverse problem that emphasizes the existence, uniqueness, and efficiency of the results. Currently, the primary obstacle to the development of CNN is the requirement of an extensive amount of data, often unavailable in real world applications, for training to achieve the best results. To this end, our previous efforts [8, 9] have developed a new topology design procedure to generate optimal structures using CNN architecture trained with large data sets generated by a Generative Adversarial Network (GAN). The discriminator in the GAN as well as the CNN were initially trained through the dataset of 3024 true optimized planar structure images generated from a conventional topology design approach SIMP. The discriminator maps the optimized structure to the key topology design parameters such as volume fraction, penalty and radius of the smoothening filter. Once the GAN is trained, the generator produced a large number of new unseen planar structures satisfying the design requirements and boundary conditions. This work has been extended to accommodate the design requirements and/or constraints in our recent effort [10] based on conditional Wasserstein generative adversarial networks (CWGAN). With CWGANs, the topology optimization conditions can be set to a required value before generating samples. CWGAN truncates the global design space by introducing an equality constraint by the designer. However, mode collapse was observed in the CWGAN training process which is responsible for the lack of diversity in the generated output structures as discussed in [10].

The purpose of this paper is to present a new constrained topology optimization approach based on a new conditional generative adversarial network (crCGAN) to not only replicate the conventional topology optimization algorithms in an extremely effective and efficient way but also completely resolve convergence instability and mode collapse problems of GAN training. The idea is to introduce a third term in the new objective function creating an inconsistency of the training samples and the constraints which in turn streamlines the training process. As previously stated, the crCGAN was created to improve convergence speed, stability, and accuracy of the topology design process while maintaining the simplicity. In addition, a minibatch discrimination technique proposed in [11] was applied in the crCGAN training process to stabilize the convergence and avoid the mode collapse. The approach has been validated by generating planar structures constrained by a desired volume ratio with the topology planar structures produced from the conventional algorithm SIMP. Additional experiments were conducted on the state-of-the art MNIST digits and CIFAR-10 images conditioned by class label. The work presented in this paper will provide a new conditional GAN technique that can enhance the convergence of the CGANs and providing significant accuracy and uniqueness of the results. The aim of this work, as well as our previous efforts [8-10], is to demonstrate a proof of concept and further scale research efforts to amalgamating deep learning with topology design optimization as well as explore a new direction in general design optimization processes.

## 2 Topology Optimization Problem Statement

Topology optimization is a modern optimization algorithm which aims to distribute material inside a given design domain [5]. Concepts of finite element method along with optimizing techniques like genetic algorithm, method of moving asymptotes, optimality criteria and level set method are utilized in this algorithm. Mathematically, it is an integer optimization problem where each finite element of the discretized design domain constitute a design variable. This design variable represents the density of that element which takes a value of either 0 (with no material) or 1 (with material), hence named a binary density variable at times. This problem can be relaxed by using converting the discrete integer problem into a continuous variable problem SIMP. In SIMP, non-binary solutions are penalized with a factor $p$. Mathematically, the topology optimization implementation suing SIMP can be defined as:

$$\min_{x}: \quad c(x) = U^T K U = \sum_{e=1}^{N}(x_e)^p \ u_e k_o u_e \tag{1}$$

$$subjected\ to: \quad \frac{V(x)}{V_0} = f,$$

$$KU = F,$$

$$0 < x_{min} \leq x \leq 1$$

where c(x) is the objective function, U and F are the global displacement and force vectors respectively. K is the global stiffness matrix, $u_e$ and $k_o$ are the element displacement vector and stiffness matrix, respectively. x is the vector of design variables. $x_{\min}$ is a vector of minimum relative densities (non-zero to avoid singularity). N (= nelx×nely) $N = \nabla_x \times \nabla_y$ is the number of elements used to discretize the design domain, p is the penalization power (typically p = 3). V (x) and $V_0$ is the material volume and design domain volume, respectively. Finally, f (volfrac) is the prescribed volume fraction. This existing method was used to generate a dataset of 3024 samples.

## 3    Related Work

### 3.1 A Novel Integrated Deep Learning Network [8]

A novel topology optimization approach was developed based on an integrated generative model consists of a GAN and a CNN in our previous work [8]. The conceptual flow of the proposed method is shown in Figure 1. In the proposed model, the discriminator is first trained through the dataset of true images followed by generated images from the generator. It is different from conventional GANs in their objective function. WGAN uses earth mover distance (EM) instead of Jenson-Shannon Divergence in GANs. In [8], WGAN was coupled with a convolutional neural network. The convolutional network was also trained using the dataset of 6048 samples. These 6048 samples are generated from existing dataset of 3024 samples using data augmentation techniques. The convolutional network maps the optimized structure to the volume fraction, penalty and radius of the smoothening filter. Once the WGAN is trained, the generator produced new unseen structures. The corresponding input variables of these new structures can be evaluated using the trained convolutional network.

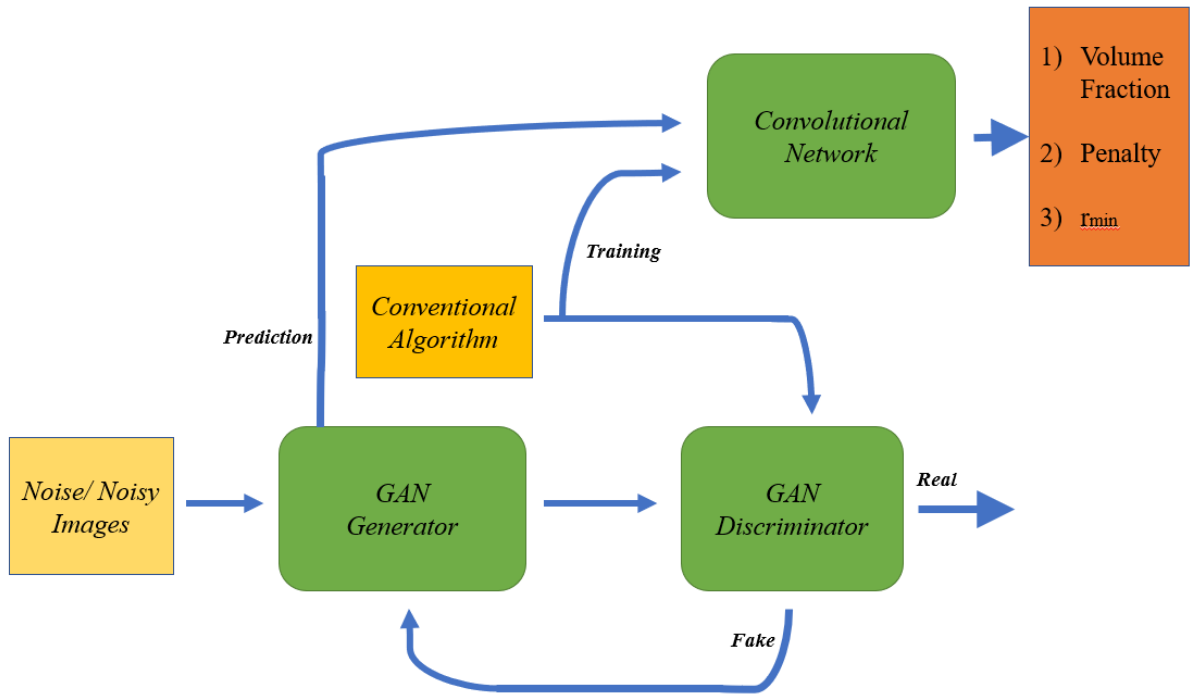

Figure 1. Conceptual flow of the proposed topology optimization framework [8]

The GAN in Figure 1 consists of 2 networks, a generator G(z) and a discriminator D(x). These two networks are adversaries of each other during training process.    The discriminator is trained to successfully distinguish between the true and fake images. Whereas, the generator is trained to generate realistic image from noise such that the discriminator is unable to distinguish between the real and fake images. Mathematically, the objective function is described as

$$\min_G \max_D \ V(D,G) = E_{x \sim P_{data}}[\log(D(x))] \ + \ E_{x \sim P_z}[\log(1 - D(G(z)))] \qquad (2)$$

where x is the image from training dataset $P_{data}$ and z is the noise vector sampled from $P_z$. There have been certain modifications to the vanilla GAN algorithm to produce good quality images. Firstly, a deep convolutional GAN (DCGAN) [12] has proved to be producing realistic images. In an attempt to stabilize the training, Wasserstein GAN (WGAN) is used where the objective function is modified and the weights are clipped to enforce the Lipchitz constraint.

## 3.2 Dataset

The training data was computed using the conventional topology optimization method, the SIMP model. A dataset of 3024 samples was generated using volume fraction (vol frac), penalty (penal) and radius of the smoothening filter ($r_{min}$) as design variables. Following is bounds of each variable:

| 1) vol_frac : 0.3 - 0.8 | 2) Penal : 2 - 4 | 3) $r_{min}$ :1.5 - 3 |
|---|---|---|

There can be more variables to this problem but since this article is a proof of concept, only 3 scalar variables are chosen. Each dataset was created at a resolution of 120 X 120 pixels with same boundary condition of a cantilever beam loaded at the mid-point of hanging end. The dataset generated was used to train the GAN and train the convolutional network. Since for training of the convolutional network, this dataset was insufficient, the dataset of 3024 samples was, hence, augmented to increase the number of samples to 6048. Augmentation of the data samples was done by adding noise at random locations for each sample. Hence, the data has 2*3024 samples for training of convolutional networks.

## 3.2 Validations

In [8], generative adversarial networks were used to construct optimal design structures for a given boundary condition and optimization settings. The SIMP model was used to generate a small number of datasets as displayed by Figure 2 left column. Samples taken from the datasets were supplied to the discriminator network in GAN for training. An adversarial game between the discriminator and the generator is performed which aids the generator to learn the distribution of dataset $P_{data}(x)$.

The convolutional neural network (CNN) in Figure 1 was trained on the same dataset from conventional algorithm. This CNN maps the optimized structure to its corresponding optimization settings. After the completion of training of GAN, samples from GAN were fed to this CNN to obtain the optimization settings. Figure 2 right column shows the samples generated from GAN after training.  It can be seen that although, some of the structures generated by GAN are not defined clearly, GANs have been able to replicate the distribution of dataset ($P_{data}$). Since there are a mix of volume fractions, it can be said that this GAN is not subjected to mode collapse.   To validate the performance of GAN, a qualitative evaluation was conducted comparing the structures from GAN against the structures from conventional algorithm with the same optimizations setting. Figure 3 presents the comparative results. It can be observed that the structures generated from GAN are extremely close to their corresponding structures from the conventional algorithm

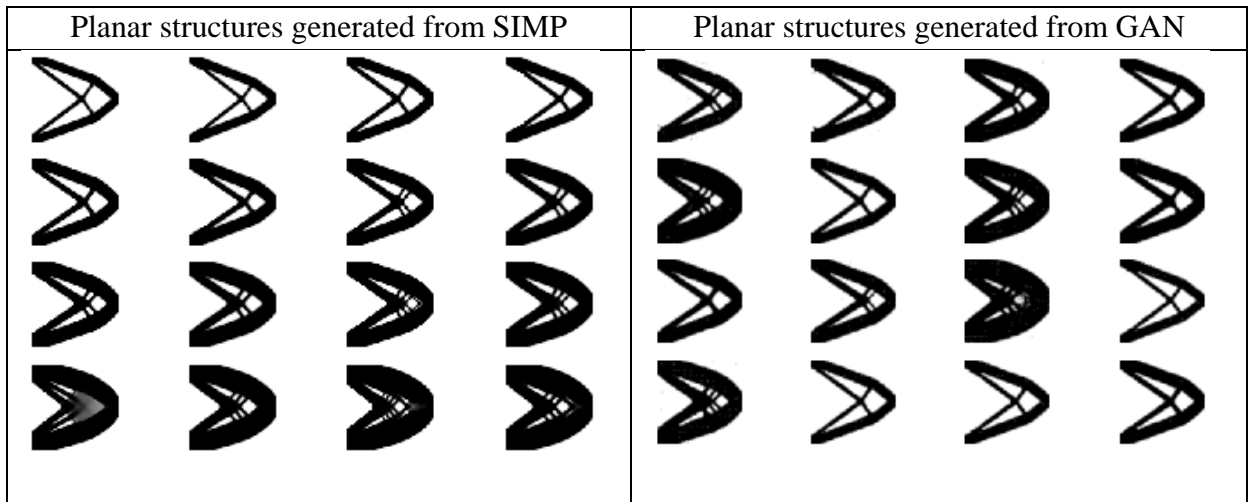

Figure 2. Optimal structures generated from SIMP (left) and GAN [8]

Planar structures from GANs are post-processed in order further improve the quality of the images. A threshold filter is created where all the pixels above a value of 0.5 are rounded to 1 and pixel values below 0.5 are rounded to 0. This filter improves the quality of the structure but the resulting structure is extremely sharp. To further smoothen the planar structure, a Gaussian filter with a kernel of 5 was applied on the structure. The resulting structure is in good agreement as can be observed in Figure 3. In summary, the proposed generative model based on GANs show a capability of generating a sub-optimal topology once trained with a very small dataset of 3024 samples. With the good agreement of the GAN results against the conventional algorithm, GANs can reconstruct models for further complex model and optimization settings which are computationally expensive.

| Generated Structure from GAN | Design parameters from Convolutional Network | Topology optimized structures from SIMP |
|---|---|---|
| 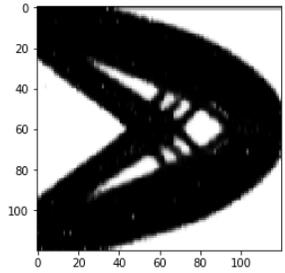 | vol_frac = 0.607<br><br>Penal = 3.0857<br><br>r$_{min}$ = 1.769 | 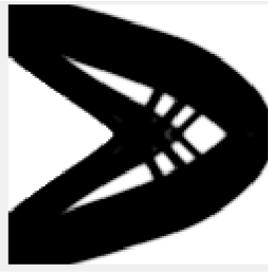 |
| 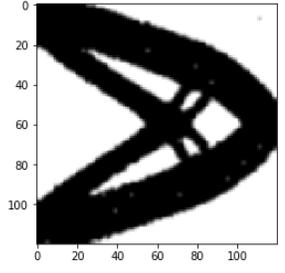 | vol_frac = 0.551<br><br>Penal = 3.607<br><br>r$_{min}$ = 2.592 | 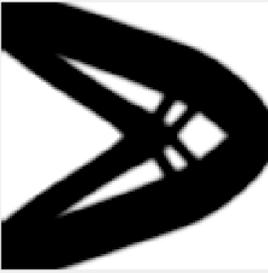 |
| 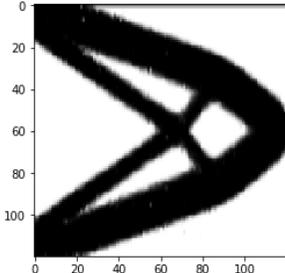 | vol_frac = 0.451<br><br>Penal = 2.861<br><br>r$_{min}$ = 1.699 | 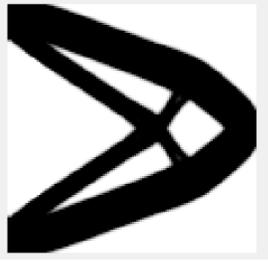 |

Figure 3. Comparative results from GAN and the corresponding conventional algorithm [8]

## 4 Constrained Topology Design Optimization by Conditional GAN

### 4.1 Conditional Generative Adversarial Nets

A typical CGAN is an extension of a GAN with required information conditioned to both the discriminator and generator. The objective function of CGANs is often described as

$$\min_G \max_D V(D,G) = E_{x \sim P_{data}}[\log(D(x|y)] + E_{z \sim P_z}[\log(1 - D(G(z|y)))] \quad (3)$$

where $x$ is the image from training dataset $P_{data}$, $z$ is the noise vector sampled from $P_z$, and $y$ is the conditioned information.

We have conducted an experiment to evaluate the planar structures generated from CGAN with desired volume density of 0.5 ($y$=0.5). The results, as presented in Figure 4, demonstrates incomparable with the planar structures produced from the conventional topology design algorithm

SIMP as shown in Figure 5 (right column). Furthermore, the results clearly show that conventional CGAN generates inconsistent volume fraction as well as poor quality planar structures. In this work, we propose a new CGAN formulation namely, convex relaxing CGAN (crCGAN) to improve the training convergence and stability in the topology design process.

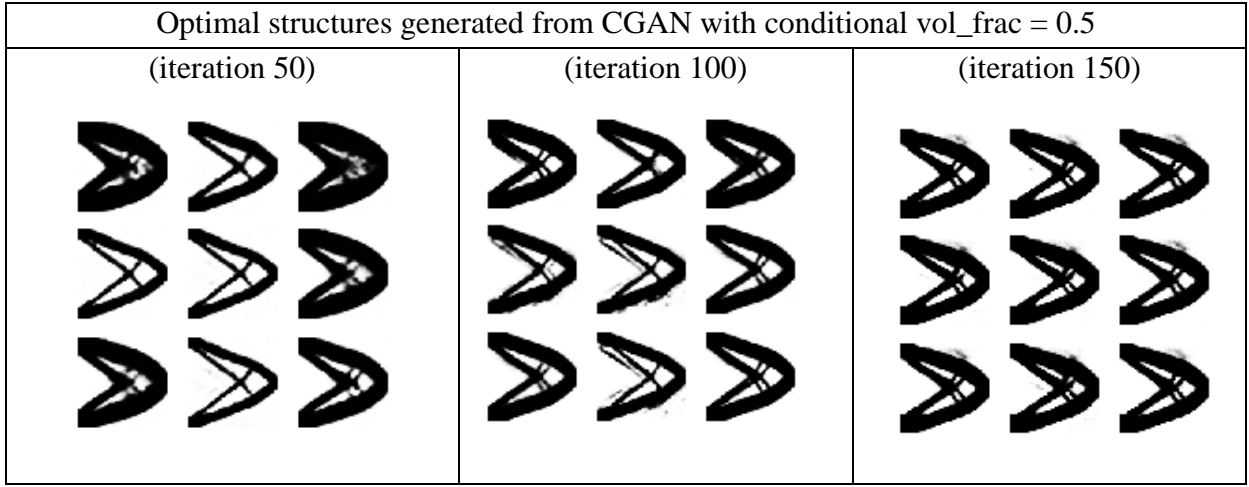

Figure 4. Planar structures generated from conventional CGAN

## 4.2 A New CGAN Formulation (crGAN)

The condition or side information in the CGAN can be considered as an equality constraint that truncates the convex-concave design space of GAN into a non-convex lower dimensional design space consisting local minima. A variety of the crCGAN formulation, as presented in Eq. (4) and Eq. (5) are introduced to relax the non-convex design space and to reform a convex or quasi-convex design space. The idea is to introduce a third term in the objective function, Eq. (4), to impose an inconsistency between the training samples and the side condition or the design constraint. The discriminator $D$ receives inputs not just training samples $x$ and the corresponding condition $y_1$ but also a random condition $y_2$. Consequently, a high order dimensional design space will be reconstructed with relaxing convexity. The same experiment was performed to evaluate planar structures generated from crCGAN with a desired volume density of 0.5 ($y_1$=0.5). The results presented in Figure 5 demonstrates very good agreement with the planar structures produced from the conventional algorithm SIMP. The variability across the generated planar structures has been significantly diluted with reasonable image quality.

$$\max_{D} \left\{ E_{x \sim P_{data}}[[\log(D(x|y_1))]] + E_{x \sim P_{data}, y_2 \sim P_y}[log[1 - D(x|y_2)]] \right.$$
$$\left. + E_{z \sim P_z}[log[1 - D(G(z|y_1))]] \right\} \quad (4)$$

$$\min_{G} \left\{ + E_{z \sim P_z}[log[1 - D(G(z|y_1))]] \right\} \quad (5)$$

Alternative crCGAN formulation Eq. (6) and Eq. (7) is also proposed in which the discriminator $D$ receives inputs from two different training samples $x_1$ and $x_2$. Verification result indicates similar

$$\max_{D} \left\{ E_{x \sim P_{data1}}[[\log(D(x_1|y))]] + E_{x \sim P_{data2}}[\log[1 - D(x_2|y)]] \right. \\ \left. + E_{z \sim P_z}[\log[1 - D(G(z|y))]] \right\} \quad (6)$$

$$\min_{G} \{ + E_{z \sim P_z}[\log[1 - D(G(z|y))]]\} \quad (7)$$

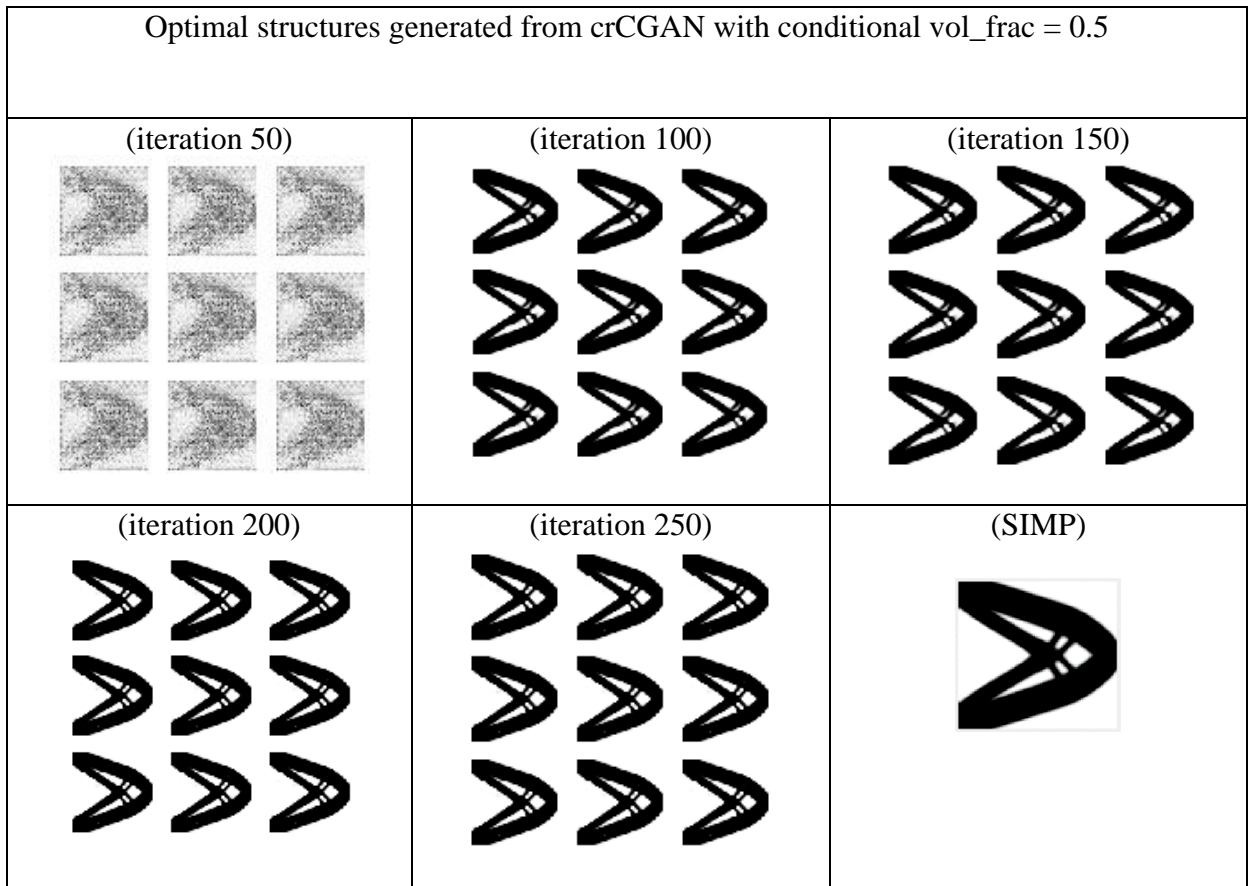

Figure 5. Planar structures generated from proposed crCGAN

## 5  Additional Experiments

We have performed additional experiments on the state-of-the art MNIST digits and CIFAR-10 images conditioned by a class label.

## 5.1 MNIST

The MNIST dataset includes 10 hand-written digits (0-9), where each digit has 5000 samples and a total of 50,000 training samples are in whole dataset. Approximately 60,000 labeled MNIST digits samples were used to train the proposed crCGAN. However, a minibatch size of only 100 samples was used to approximate the gradient for each training step. One training step consists of one update of discriminator (D) followed by one update of generator (G). One iteration requires 500 training steps to go through all the training samples. The results shown in Figure 6 are the generated images at iteration 88 and iteration 186 by the original CGAN and the proposed crCGAN, respectively. By the introducing the third term, as stated before, an inconsistency between the training samples and the side condition was imposed in the design space so that the discriminator can easily distinguish the difference between real and fake images as well as match the corresponding classes. In turn, the generator was able to learn the data distribution of a particular class from the feedback of the discriminator. The training data sets were setting up with 50, 100, and 150 labeled samples. the generator was able to quickly learn the data distribution of a particular class from the feedback of the discriminator.

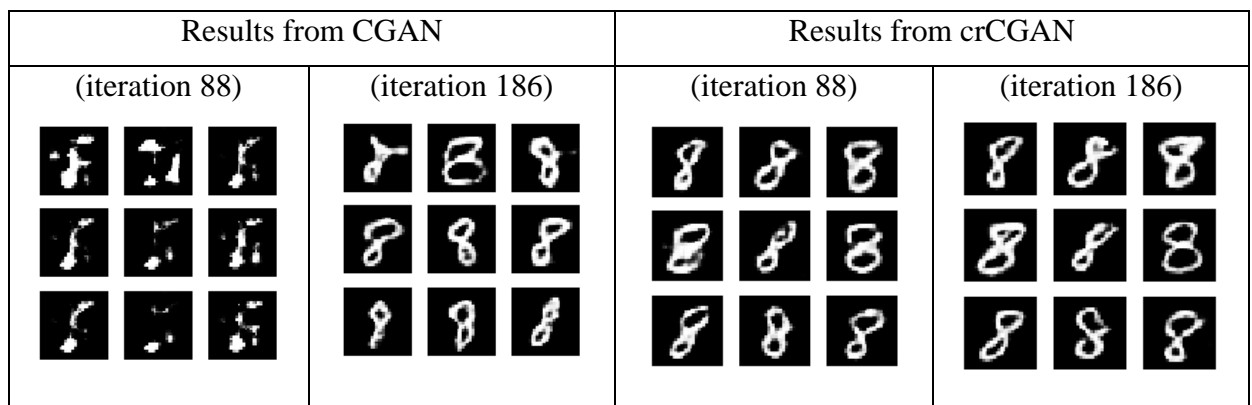

Figure 6. Samples generated from traditional CGAN (left) and crCGAN (right)

## 5.2 CIFAR-10

The training procedure applied to the MNIST data was also applied on the CIFAR-10 dataset. The CIFAR-10 also has 10 classes of images. Each class has 5000 images and a total of 50000 labeled samples are in the dataset. For each iteration, the generator (G) generates a 10*10 image matrix, where each column contains the image from same class (same label). Figure 7 shows the results at iteration 150 and iteration 300 for the original CGAN (left column) and our proposed crCGAN (right column). With side information provided by the third term, the crCGAN was able to quickly learn the conditioned distribution of each class and generate meaningful images.

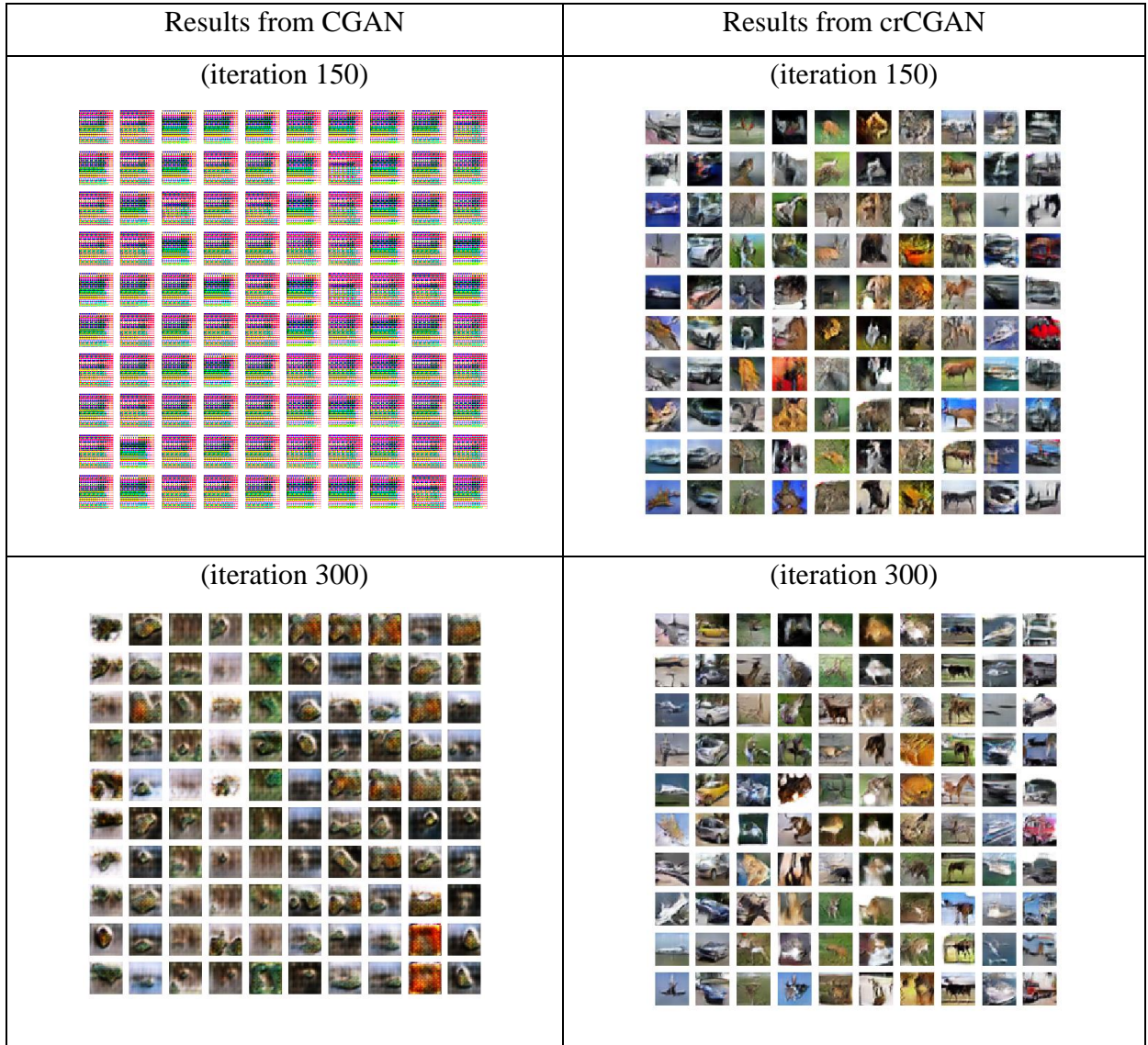

Figure 7. Samples generated from traditional CGAN (left) and crCGAN (right)

## 6 Improved crCGA Training using Minibatch Discrimination

As reported in recent applications of GANs [11], we have suffered similar mode collapse problem in both traditional CGAN and crCGAN training processes, even though crCGAN shown much stable and rapidly convergence. The discussion and explanation of the mode collapse and non-convergence problems of the GANs is well presented in [11]. Probably the most effective improved technique for avoiding mode failures is the minibatch discrimination. The concept of the minibatch discrimination is, as stated in [11], *"any discriminator model that looks at multiple examples in combination, rather than in isolation, could potentially help avoid collapse the generator"*. We have applied the minibatch technique on our crCGAN with CIFAR-10 data sets. As shown in Figure 8 (left), the generator of crCGAN failed at 635 iteration and feature appearing does not regained in 727 iteration. With minibatch discrimination in crCGAN training, the mode collapse vanish and the image quality singingly in all iterations as shown in Figure 8 (right).

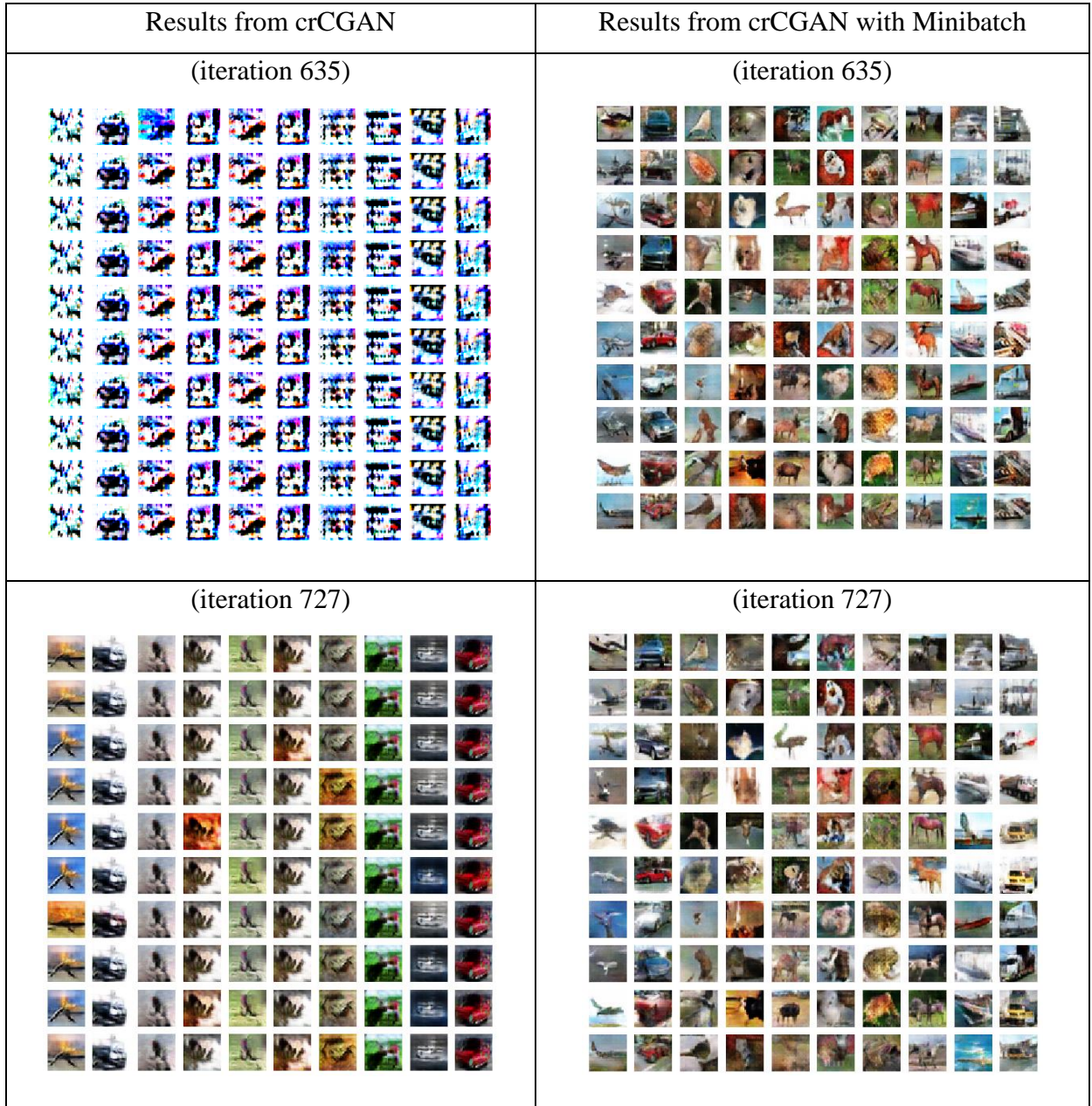

Figure 8. Samples generated from crCGAN (left) and crCGAN with Minibatch (right)

# 7 Conclusions

To address the shortcomings of conventional topology design approaches, this paper presents a shift in the fundamental approach to the topology design paradigm by using crCGAN, a new CGAN architecture, to optimize the initial design topology effectively and rapidly. In addition, a minibatch discrimination technique was applied in the crCGAN training to ensure stable convergence and avoiding mode collapse. This paper presents follow up work of our on-going effort toward a new frontier in deep learning based structural topology design optimization.

**Acknowledgment**

The authors thank the Ohio Supercomputer Center for providing High-Performance Computing resources and expertise to the authors. OSC is a member of the Ohio Technology Consortium, a division of the Ohio Department of Higher Education.